\documentclass[conference,a4paper]{IEEEtran}

\IEEEoverridecommandlockouts
\usepackage{cmap}
\usepackage[T1]{fontenc}
\usepackage[utf8]{inputenc}
\usepackage{cite}
\usepackage{amsmath}
\usepackage{amssymb}
\usepackage{array}
\usepackage{balance}
\usepackage{graphicx}
\usepackage{textcomp}
\usepackage{url}

\newcommand{\TUDUM}{T\"UD\"UM}
\newcommand{\thinkopen}{\texttt{\textless think\textgreater}}
\newcommand{\thinkclose}{\texttt{\textless/think\textgreater}}

\begin{document}

\title{\TUDUM: A Turkish-Thinking Reasoning Pipeline for Qwen3.5-27B}

\author{
\IEEEauthorblockN{Baran Bing\"ol}
\IEEEauthorblockA{Department of Artificial Intelligence and Data Engineering\\
Faculty of Engineering\\
Ankara University\\
Ankara, T\"urkiye\\
22290007@ogrenci.ankara.edu.tr}
\and
\IEEEauthorblockN{Bahaeddin T\"urko\u{g}lu}
\IEEEauthorblockA{Department of Artificial Intelligence and Data Engineering\\
Faculty of Engineering\\
Ankara University\\
Ankara, T\"urkiye\\
turkoglub@ankara.edu.tr}
}

\maketitle

\begin{abstract}
This paper presents \TUDUM{} (\emph{T\"urk\c{c}e D\"u\c{s}\"unen \"Uretken Model}), a project pipeline for adapting a Qwen-family 27B thinking model toward Turkish reasoning. The central problem is not only to answer Turkish prompts in Turkish, but to make the explicit reasoning trace itself Turkish. A thinking model may translate a Turkish prompt into an English-centered internal or visible scratchpad, solve the problem mostly in English, and only localize the final answer. \TUDUM{} instead treats the generated \thinkopen{}...\thinkclose{} block as a trainable behavior. The pipeline starts from the project base checkpoint \texttt{unsloth/Qwen3.5-27B}, applies supervised fine-tuning (SFT) on 15,991 Turkish reasoning examples using LoRA adapters, and then applies GRPO-family reinforcement learning on a proxy-filtered Turkish mathematics environment. The results are mixed. SFT made the model shorter and more consistently Turkish in its reasoning behavior, with large reductions in average response length and thinking exhaustion, but reduced benchmark accuracy. RL recovered some mathematical performance, especially AIME24 at the best early checkpoint, yet did not uniformly improve all benchmarks and did not exceed the base model on the reported Macro-6 average. The contribution is therefore best framed as a technically honest Turkish-thinking reasoning pipeline and evaluation, not as a claim of state-of-the-art Turkish reasoning. The released step-50 model is publicly available.
\end{abstract}

\vspace{-0.25em}
\begin{center}
\scriptsize
\raisebox{-0.35em}{\includegraphics[height=1.35em]{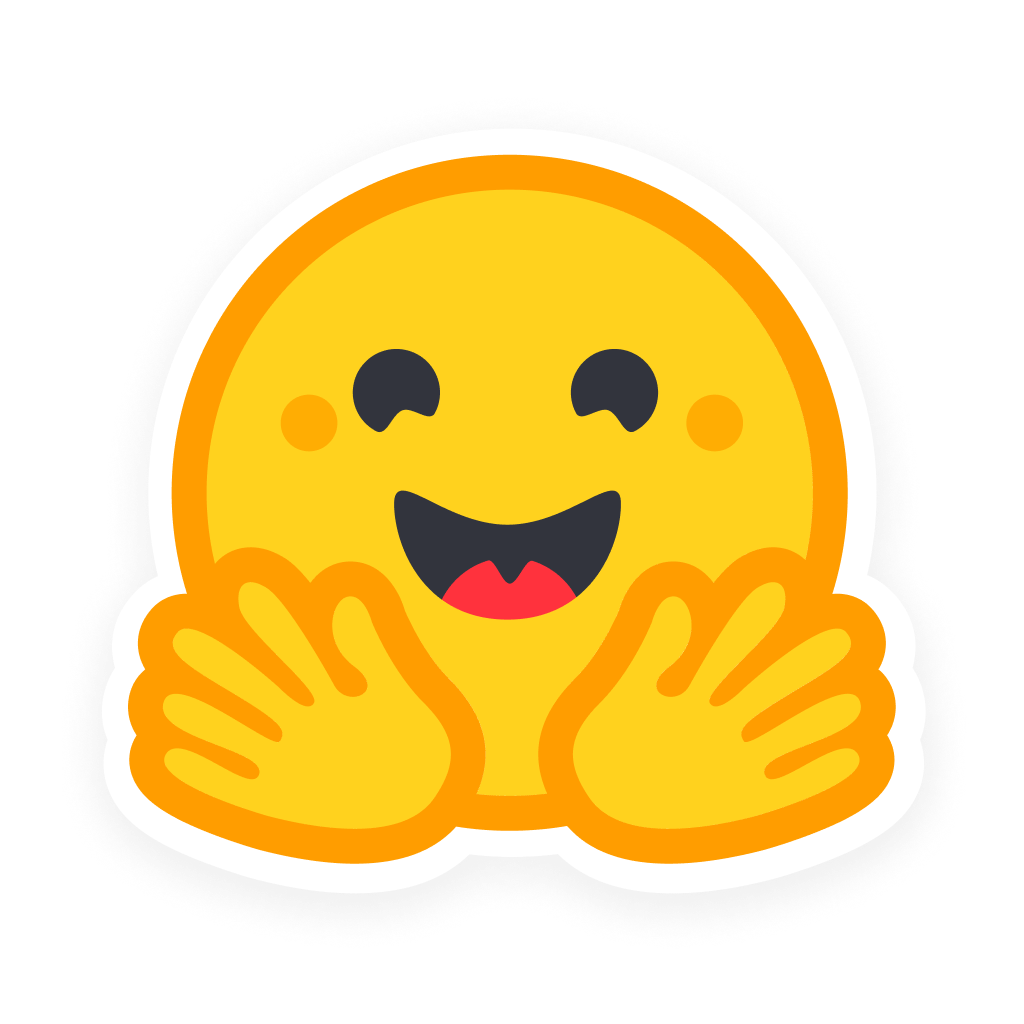}}\hspace{0.35em}
\textbf{Model availability:} \url{https://huggingface.co/barandinho/qwen3.5-27b-tudum-dapo-50}
\end{center}
\vspace{-0.65em}


\section{Introduction}
Reasoning models expose intermediate text before the final answer. In the \TUDUM{} project this behavior is represented as a thinking block followed by a final response:
\begin{quote}
\small
\texttt{\textless think\textgreater}\\
\texttt{intermediate reasoning trace}\\
\texttt{\textless/think\textgreater}\\
\texttt{final answer}
\end{quote}
This output format changes the multilingual problem. A model can receive a Turkish prompt and produce a Turkish final answer while still using English for most of the reasoning trace. This behavior may be acceptable for making the answer fit the local language. But this is not reasoning in Turkish. This difference is important for education, for checking how the answer was produced, and for later use in Turkish-language workflows. Because the user or evaluator does not only see the final answer. They also see the order of the mathematical, scientific, or coding decisions that led to that answer.

The project base checkpoint is \texttt{unsloth/Qwen3.5-27B}. Its model card lists \texttt{Qwen/Qwen3.5-27B} as the base model, and the official Qwen3.5 model card states that Qwen3.5 models operate in thinking mode by default \cite{qwen35hf,unslothqwen35}. For broader Qwen-family background, we also cite the Qwen3 technical report \cite{yang2025qwen3}. Prior work on chain-of-thought prompting showed that explicit intermediate steps can improve complex reasoning in large language models \cite{wei2022cot}. Multilingual chain-of-thought work further showed that reasoning can transfer across languages at sufficient scale \cite{shi2022mgsm}. However, newer analyses also suggest that multilingual models can use English-centered internal representations even when both input and output are non-English \cite{schut2025english}. \TUDUM{} is motivated by this gap: Turkish output is not sufficient evidence that the model reasons in Turkish.

This paper summarizes the \TUDUM{} pipeline and its measured behavior on the project evaluation suite. The exact revision hashes of the base model, tokenizer, chat template, and evaluation harness are not included in the supplied project files and remain reproducibility items.

The contributions are:
\begin{itemize}
\item a Turkish-thinking adaptation pipeline consisting of Turkish reasoning SFT followed by GRPO-family RL;
\item a clear separation between answer language and reasoning-trace language;
\item an empirical comparison of Base, SFT, and RL checkpoints across math, knowledge, science, code, and instruction-following benchmarks;
\item an explicit accounting of limitations and future RL environments needed to recover regressions.
\end{itemize}

\section{Method}

\subsection{Foundation Model and Thinking Format}
The foundation in the supplied project report is \texttt{unsloth/Qwen3.5-27B}, whose model card identifies \texttt{Qwen/Qwen3.5-27B} as its base. We do not restate architectural details that are absent from the supplied project files. Operationally, the model is used as a thinking model: the assistant response begins at \thinkopen{}, continues with a reasoning trace, closes with \thinkclose{}, and then emits the final answer. The Qwen3.5 model card states that thinking mode is default and that Qwen3.5 does not officially support the Qwen3 \texttt{/think} and \texttt{/nothink} soft switch; the \TUDUM{} setup therefore treats \thinkopen{}...\thinkclose{} as the observed structured response format rather than as a separately validated switch mechanism \cite{qwen35hf}. The training target is not just the final answer. It is a structured response. The first part of this response is expected to include reasoning in Turkish.

\begin{figure}[t]
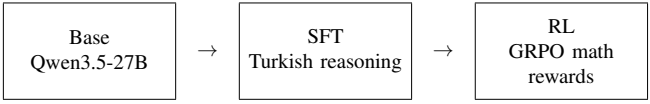

\centering
\resizebox{\columnwidth}{!}{%
\begin{tabular}{c@{\quad$\rightarrow$\quad}c@{\quad$\rightarrow$\quad}c}
\fbox{\parbox[c][0.55in][c]{1.05in}{\centering Base\\Qwen3.5-27B}} &
\fbox{\parbox[c][0.55in][c]{1.05in}{\centering SFT\\Turkish reasoning}} &
\fbox{\parbox[c][0.55in][c]{1.05in}{\centering RL\\GRPO math rewards}}
\end{tabular}}
\caption{The \TUDUM{} training pipeline. The contribution is the full adaptation and evaluation pipeline rather than a standalone dataset.}
\label{fig:pipeline}
\end{figure}

This framing is important because the model can otherwise satisfy a Turkish final-answer requirement by translating only the last answer. In \TUDUM{}, the behavior being trained is that the scratchpad-like text inside the thinking block stays Turkish unless mathematical notation, code, or symbols make language classification irrelevant. Fig.~\ref{fig:pipeline} summarizes the resulting Base--SFT--RL pipeline.

\subsection{Supervised Fine-Tuning}
Supervised fine-tuning (SFT) trains the model to imitate target completions by maximizing the likelihood of desired assistant responses, as in instruction-tuning pipelines that begin with supervised demonstrations before preference or RL stages \cite{ouyang2022instructgpt}. In this project, SFT was used to teach a response style: Turkish reasoning inside the thinking block followed by a final answer. This is different from training only a Turkish answer formatter. For a Turkish prompt $x$ and target assistant tokens $y_{1:T}$, the masked SFT objective can be written as
\begin{equation}
\mathcal{L}_{\mathrm{SFT}}(\theta)
= - \sum_{t=1}^{T} m_t \log p_\theta(y_t \mid x, y_{<t}),
\label{eq:sft}
\end{equation}
where $m_t=1$ only for assistant-response tokens in the supervised region and $m_t=0$ for untrained prompt tokens. In the reported setup, this supervised region begins at \thinkopen{}, so the loss directly covers the reasoning trace.

The final SFT dataset contains 15,991 examples. Its composition is shown in Table~\ref{tab:sftdata}. Mathematics dominates the dataset, but science, code, and system-prompt examples were deliberately included only in SFT to broaden the behavior relative to the RL stage.

\begin{table}[t]
\caption{Final SFT dataset composition. Values are from the supplied experiment document.}
\label{tab:sftdata}
\centering
\footnotesize
\begin{tabular}{lrr}
\hline
Component & Rows & Share \\
\hline
Mathematics & 9,566 & 59.8 \\
Science & 5,134 & 32.1 \\
Code & 652 & 4.1 \\
System prompts & 639 & 4.0 \\
\hline
Total & 15,991 & 100.0 \\
\hline
\end{tabular}
\end{table}

The mathematics subset was generated from the public \texttt{barandinho/DAPO-Math-14k-Turkish} dataset \cite{barandinho2026dapomathtr} using a DeepSeek-V3.2 teacher \cite{deepseek2025v32}, filtered for prompt quality, Turkish reasoning language, and \texttt{math\_verify} correctness \cite{kydlicek2026mathverify}. The reported flow was 12,368 successful generations, 10,876 Turkish-reasoning candidates, 9,574 correct examples, and 9,566 final deduplicated rows. The science subset was generated from Turkish science prompts and judged with Kimi-K2.5 \cite{kimi2026k25}, yielding 5,134 final rows. The code subset similarly used the project teacher generation path and Kimi-K2.5 correctness filtering, yielding 652 final rows. These teacher and judge details are reported here only as project methodology, not as a claim that the resulting data is free of judge noise.

\subsection{LoRA Adaptation}
The SFT stage used LoRA rather than full-parameter fine-tuning. LoRA freezes the original large weight matrices and trains low-rank adapter matrices whose product supplies the update \cite{hu2021lora}. For a base matrix $W\in\mathbb{R}^{d_{\mathrm{out}}\times d_{\mathrm{in}}}$, the adapted matrix is
\begin{equation}
W' = W + \Delta W,\qquad
\Delta W = \frac{\alpha}{r}BA,
\label{eq:lora}
\end{equation}
where $A\in\mathbb{R}^{r\times d_{\mathrm{in}}}$ and $B\in\mathbb{R}^{d_{\mathrm{out}}\times r}$ are trained while $W$ remains frozen. This greatly reduces the number of trainable parameters compared with updating the entire 27B model. The supplied report specifies LoRA rank $r=64$, alpha $\alpha=64$, and dropout 0. The SFT setup used 4 H200 GPUs, Unsloth with TRL SFTTrainer, maximum sequence length 32,768, five epochs, learning rate $2\times10^{-4}$ with cosine schedule, and effective batch size 16.

\subsection{GRPO-Family Reinforcement Learning}
The RL stage did not reuse the full SFT mixture. It used a separate mathematics environment derived from the project DAPO-Math Turkish 5k dataset. A Qwen3.5-4B proxy model generated 12 completions per question. Each completion was checked against the gold answer with \texttt{math\_verify}. For prompt $i$, the proxy pass rate was
\begin{equation}
\mathrm{ppr}_i =
\frac{1}{12}\sum_{k=1}^{12}
\mathbf{1}\{\mathrm{verify}(c_{i,k}, a_i)=1\}.
\label{eq:ppr}
\end{equation}
Questions with proxy pass rate 0 or 1 were removed; only $0 < \mathrm{ppr} < 1$ remained. This removed 1,652 always-failed and 531 always-solved questions, leaving 2,339 RL training prompts.

The RL algorithm is best described as GRPO with DAPO-style training stabilizers and GDPO-style multi-reward normalization. GRPO, introduced in DeepSeekMath, samples a group of completions for the same prompt, scores them, and uses group-relative rewards as the baseline instead of training a separate value model \cite{shao2024deepseekmath}. For a group of $G$ completions with scalar rewards $R_1,\ldots,R_G$, the project-level explanation of GRPO advantages is
\begin{equation}
A_j = \frac{R_j-\mu_G}{\sigma_G+\epsilon},\quad
\mu_G=\frac{1}{G}\sum_{j=1}^{G}R_j,
\label{eq:advantage}
\end{equation}
where $\sigma_G$ is the within-group reward standard deviation. The clipped policy update is related to PPO-style optimization \cite{schulman2017ppo}. With DAPO clip-higher, the upper clipping bound is relaxed; a simplified sequence-level form is
\begin{equation}
\begin{split}
\mathcal{L}_{\mathrm{RL}} =
&-\frac{1}{G}\sum_{j=1}^{G}
\min\!\bigl(\rho_j A_j,\\
&\operatorname{clip}(\rho_j,1-\epsilon,1+\epsilon_h)A_j\bigr)
+\beta\,\mathrm{KL}(\pi_\theta\Vert\pi_{\mathrm{ref}}),
\end{split}
\label{eq:rl}
\end{equation}
The project notes further identify DAPO-style settings: asymmetric clipping, dynamic sampling, token-level/DAPO loss normalization, overlong filtering, and soft overlong penalties \cite{yu2025dapo}. Because the run used multiple reward heads, the notes also identify \texttt{scale\_rewards=gdpo}, corresponding to reward-decoupled normalization for multi-reward RL \cite{liu2026gdpo}.

Table~\ref{tab:reward} summarizes the active RL recipe. Reward design matters because the model optimizes the reward actually supplied, not the informal goal. In \TUDUM{}, correctness is the main reward component. Format, Turkish-language behavior, and length control are used as supporting signals. The scalar reward used by the reported run can be summarized as
\begin{equation}
R = 1.0R_{\mathrm{acc}} + 0.25R_{\mathrm{fmt}}
+0.5R_{\mathrm{tr}} + 0.2R_{\mathrm{len}},
\label{eq:reward}
\end{equation}
These components represent mathematical correctness, correct format, Turkish-language consistency, and soft control of overly long answers. The language reward is designed to make both the final answer and the reasoning language more Turkish. However, its weight can create a trade-off. If the language reward is too strong, it may compete with correctness or instruction following.
\begin{table}[t]
\caption{GRPO/RL summary for the reported run. Values are from the supplied GRPO notes and experiment document.}
\label{tab:reward}
\centering
\scriptsize
\begin{tabular}{@{}p{0.25\columnwidth}p{0.24\columnwidth}p{0.30\columnwidth}@{}}
\hline
Item & Value & Purpose \\
\hline
Core objective & \texttt{rlhf\_type=grpo} & Group-relative policy update without a critic \\
Loss variant & \texttt{loss\_type=dapo} & DAPO-style token-level normalization \\
Group size & 8 generations & Completions compared per prompt \\
KL coefficient & $\beta=0.01$ & Reference-policy regularization \\
Reward scaling & \texttt{gdpo} & Per-reward normalization for multiple rewards \\
Clipping & $\epsilon=0.2$, $\epsilon_{\mathrm{high}}=0.28$ & DAPO clip-higher exploration \\
Sampling & dynamic, max resample 3 & Avoid zero-variance groups \\
Length handling & soft max 16,384; cache 8,192 & Overlong filtering and penalty \\
\hline
\texttt{math\_accuracy} & weight 1.0 & Gold-answer correctness via math verifier \\
\texttt{math\_format} & weight 0.25 & Exactly one \thinkclose{} \\
\texttt{lang\_turkish} & weight 0.5 & Turkish consistency and no Chinese characters \\
\texttt{soft\_overlong} & weight 0.2 & Linear length penalty in $[-1,0]$ \\
\hline
\end{tabular}
\end{table}

\section{Evaluation}
The study compared the base model, the SFT checkpoint, and the RL checkpoints at steps 50, 100, 150, 200, 250, and 300. The evaluation suite does not focus on only one type of task. It measures different abilities. AIME24 and AIME25 measure math reasoning in the style of competition problems. Turkish MMLU measures general knowledge and problem-solving ability in Turkish. It follows the MMLU-style evaluation format \cite{hendrycks2021mmlu,bayram2024turkishmmlu}. GPQA measures scientific question answering at the graduate level \cite{rein2023gpqa}. HumanEval measures whether the generated code works correctly \cite{chen2021humaneval}. IFEval measures whether the model can follow instructions with clear and checkable rules \cite{zhou2023ifeval}; In this project, both prompt-level and instruction-level IFEval scores are reported.

Macro-6 is the average score of six tests:AIME24, AIME25, Turkish MMLU, GPQA, HumanEval, and IFEval prompt:
\begin{equation}
\begin{split}
\mathrm{Macro\mbox{-}6} = \frac{1}{6}(&s_{\mathrm{A24}}+s_{\mathrm{A25}}+s_{\mathrm{MMLU}}\\
&+s_{\mathrm{GPQA}}+s_{\mathrm{HE}}+s_{\mathrm{IFp}}).
\end{split}
\label{eq:macro}
\end{equation}
The IFEval instruction score is not included in this average and is reported separately as an additional diagnostic score.

\section{Results}
Table~\ref{tab:main_eval} shows the main comparison. The RL column uses only one checkpoint: step 50. This is because step 50 has the highest Macro-6 average among the recorded RL checkpoints. The best checkpoint for each individual benchmark is not used in the main comparison. These checkpoints are reported separately in Table ~\ref{tab:rlsteps} as diagnostic evidence. The clearest result is that SFT improved the target language behavior, but it reduced benchmark accuracy. The base model has a Macro-6 score of 81.7. The SFT model has a Macro-6 score of 75.8. RL at step 50 reaches a Macro-6 score of 78.1. So, RL recovered part of the performance loss caused by SFT. However, RL did not perform better than the base model in the reported macro average.

\begin{table}[t]
\caption{Main Base, SFT, and RL comparison. Scores are accuracies in percent. RL step 50 is selected once by Macro-6.}
\label{tab:main_eval}
\centering
\footnotesize
\begin{tabular}{lrrr}
\hline
Benchmark & Base & SFT & RL step 50 \\
\hline
AIME24 & 82.2 & 78.9 & 86.7 \\
AIME25 & 74.4 & 67.8 & 72.2 \\
Turkish MMLU & 82.3 & 79.5 & 79.4 \\
GPQA & 73.1 & 71.6 & 67.8 \\
HumanEval & 89.4 & 84.1 & 87.8 \\
IFEval prompt & 88.6 & 72.9 & 74.9 \\
IFEval instruction & 91.8 & 79.5 & 80.5 \\
\hline
Macro-6 average & 81.7 & 75.8 & 78.1 \\
\hline
\multicolumn{4}{p{0.92\columnwidth}}{\footnotesize Macro-6 is the average score of six tests: AIME24, AIME25, Turkish MMLU, GPQA, HumanEval, and IFEval prompt. The IFEval instruction score is not included in this average and is shown separately.}
\end{tabular}
\end{table}

\begin{table*}[t]
\caption{Diagnostic RL checkpoint scores by step. Scores are accuracies in percent; the ``Best'' column is per benchmark and is not used as the main model comparison.}
\label{tab:rlsteps}
\centering
\scriptsize
\begin{tabular}{lrrrrrrl}
\hline
Benchmark & 50 & 100 & 150 & 200 & 250 & 300 & Best \\
\hline
AIME24 & 86.7 & 82.2 & 78.9 & 75.6 & 80.0 & 82.2 & 50 (86.7) \\
AIME25 & 72.2 & 68.9 & 71.1 & 62.2 & 66.7 & 72.2 & 50 (72.2) \\
Turkish MMLU & 79.4 & 79.2 & 79.3 & 79.7 & 79.5 & 79.6 & 200 (79.7) \\
GPQA & 67.8 & 70.0 & 72.3 & 70.7 & 70.9 & 70.5 & 150 (72.3) \\
HumanEval & 87.8 & 86.8 & 85.2 & 85.4 & 84.1 & 86.0 & 50 (87.8) \\
IFEval prompt & 74.9 & 75.7 & 74.6 & 74.6 & 74.9 & 74.0 & 100 (75.7) \\
IFEval instruction & 80.5 & 81.6 & 81.0 & 80.7 & 80.7 & 80.1 & 100 (81.6) \\
\hline
Macro-6 average & 78.1 & 77.1 & 76.9 & 74.7 & 76.0 & 77.4 & 50 (78.1) \\
\hline
\end{tabular}
\end{table*}

\subsection{SFT Effects}
The main effect of the SFT stage was behavioral. The provided report says that the base Qwen3.5-27B checkpoint tended to switch to English on difficult problems. In contrast, the SFT model mostly stopped this behavior. It started to produce more controlled reasoning in Turkish. In this sense, the project changed the visible behavior from Turkish final-answer localization toward Turkish visible reasoning. Quantitatively, Table~\ref{tab:length} shows large reductions in response length and thinking exhaustion. AIME24 average response length decreased from 7,163 words to 4,202 words, Turkish MMLU from 1,266 to 404 words, and IFEval from 1,752 to 330 words. Thinking exhaustion also decreased: AIME24 from 15.6\% to 2.2\%, and AIME25 from 20.0\% to 7.8\%.

\begin{table}[t]
\caption{Reasoning-length and thinking-exhaustion indicators after SFT. Values are from the supplied experiment document.}
\label{tab:length}
\centering
\footnotesize
\begin{tabular}{lrr}
\hline
Indicator & Base & SFT \\
\hline
AIME24 avg. words & 7,163 & 4,202 \\
Turkish MMLU avg. words & 1,266 & 404 \\
IFEval avg. words & 1,752 & 330 \\
AIME24 thinking exhaustion & 15.6\% & 2.2\% \\
AIME25 thinking exhaustion & 20.0\% & 7.8\% \\
\hline
\end{tabular}
\end{table}

These length changes should not be overinterpreted as purely linguistic. The teacher data was generated with controlled output budgets, including a 4,096-token post-thinking completion budget and a 28,672-token thinking budget for math and science generation. Because SFT optimized the assistant response from \thinkopen{} onward inside a 32K context, the model likely learned both Turkish reasoning style and the shorter scratchpad distribution of the teacher data.

\subsection{RL Effects}
RL improved some math behavior but not all benchmarks. Step 50 produced the strongest overall RL result and the best AIME24 result: 86.7\%, above both base and SFT on that benchmark. AIME25 also improved over SFT at step 50, from 67.8\% to 72.2\%, but remained below the base score of 74.4\%. The diagnostic checkpoint table shows that some benchmarks reached their best scores at later steps. GPQA reached its best score at step 150 with 72.3\%. Turkish MMLU reached its best score at step 200 with 79.7\%. IFEval reached its best score at step 100. These best scores for individual benchmarks are useful for understanding how training changed over time. However, they are not treated as one single deployable model. The IFEval prompt score recovered only a small part of the loss caused by SFT. The base model scored 88.6\% on IFEval prompt. The SFT model scored 72.9\%. The best diagnostic RL checkpoint reached 75.7\%.

The results therefore support a limited claim: proxy-filtered math RL can recover and sometimes improve math performance after Turkish-reasoning SFT, but a math-only RL environment does not uniformly restore broad capability or instruction following. This is consistent with the study design. The RL environment was intentionally narrower than the SFT dataset.

\section{Discussion}
\TUDUM{} should be seen as a pipeline for controlling the language of explicit reasoning in large language models. The aim of this study is not only to improve the accuracy of the final answer. It is also to make the reasoning process more stable, more transparent, and more usable in Turkish. This matters because Turkish-language education, science, and coding workflows need models that can reason in Turkish, not models that only translate the final answer into Turkish. The results show that the SFT stage mainly changed the behavior of the model. After SFT, the model became more consistent in producing reasoning traces in Turkish. This means that supervised fine-tuning can teach the model the expected reasoning format and the expected language behavior. LoRA also made this adaptation practical, because it allowed the model to be adapted without updating all 27B parameters.However, the results also show an important trade-off. SFT improved Turkish reasoning behavior, but it reduced benchmark accuracy. The RL stage recovered part of this loss, especially in mathematics. However, RL did not outperform the base model on the reported Macro-6 average. This shows that controlling the language of reasoning and improving benchmark accuracy are connected goals, but they are not the same goal.

One important reason for this result is the difference between the SFT data and the RL data. The SFT stage used examples from mathematics, science, coding, and system-prompt following. In contrast, the RL stage used only proxy-filtered mathematics data. For this reason, it is expected that RL improved some mathematics scores more than it improved broad knowledge, instruction following, or code generation. This shows that the scope of the reward data must match the scope of the target behavior. The reward design is also central to the final behavior of the model. Format and language rewards are useful because they reduce malformed outputs and help prevent English drift. However, they cannot replace task-specific verifiers. At the same time, a reward that focuses only on correctness may improve benchmark accuracy, but it may also allow weak formatting or English reasoning traces. Therefore, Turkish reasoning, correctness, instruction following, code execution, and scientific answering should be optimized together as balanced objectives.

Overall, the findings show that TÜDÜM is a useful first step toward controlled Turkish reasoning in large language models. However, the current RL stage is limited because its reward environment is based only on mathematics. Future work should extend the RL data and reward functions beyond mathematics. In particular, future versions should include verifiable tasks for instruction following, coding, scientific question answering, and broad Turkish knowledge. This would help preserve Turkish reasoning behavior while also improving general benchmark performance.

\section{Future Work}
Future work should go beyond math-only reinforcement learning. The most important next step is to create broader and verifiable training environments. These environments should include Turkish IFEval for instruction following, open-ended science and knowledge verification, code execution with partial test-case scoring, procedural logic puzzles, and selected SFT prompts to protect stability. This broader setup directly targets the main weakness observed in the results. After SFT, IFEval performance dropped sharply. Math-only RL did not recover this drop. This shows that future RL should not optimize only mathematical correctness.It should also protect instruction following, Turkish reasoning behavior, coding ability, and general knowledge performance. Future RL should also use a stronger training setup. This setup should include larger and more diverse prompt pools, careful difficulty filtering, more balanced reward weights, and more conservative hyperparameters. In particular, future experiments should avoid prompt groups that are always correct or always wrong. They should also use a lower learning rate, more generations per prompt, and less aggressive length handling. Larger models or stronger base checkpoints may improve the results. However, the current findings suggest that model size alone is not the main problem. The training environments and reward functions must represent the behaviors that the model is expected to keep. In other words, a Turkish reasoning model should not be trained and rewarded only for giving correct answers. It should also be trained and rewarded for reasoning in Turkish, following instructions, generating working code, and answering scientific questions reliably.

\section{Limitations}
The evidence in this paper is limited to the supplied project reports. The public model-card provenance is identified, but exact revision hashes, benchmark prompts, evaluation harness commit, random seeds, confidence intervals, and post-training token-ratio audits are not independently reconstructed here. The result should therefore be read as an internal project evaluation rather than a fully reproducible public benchmark submission. We also do not claim that the visible thinking trace is a faithful record of all internal computation. The claim is narrower: the model was trained and evaluated to produce Turkish explicit reasoning traces in the specified output format.

\section{Conclusion}
\TUDUM{} addresses a concrete multilingual reasoning gap: Turkish final answers are not the same as Turkish reasoning. Starting from a Qwen-family 27B thinking model, the project used LoRA SFT to teach Turkish reasoning traces and GRPO-family RL to optimize math correctness, format, Turkish consistency, and length behavior. The empirical result is mixed but informative. SFT improved the target behavior while reducing benchmark accuracy; RL recovered some math performance but did not uniformly improve the suite or beat the base Macro-6 score. The next step is broader multi-environment RL with rewards that measure the full target behavior.

\section*{Acknowledgment}
The numerical calculations reported in this paper were partially performed at TUBITAK ULAKBIM, High Performance and Grid Computing Center (TRUBA resources).

\balance

\end{document}